\documentclass{article}

\usepackage{neurips_2026}

\usepackage[utf8]{inputenc} % allow utf-8 input
\usepackage[T1]{fontenc}    % use 8-bit T1 fonts
\usepackage{hyperref}       % hyperlinks
\usepackage{url}            % simple URL typesetting
\usepackage{booktabs}       % professional-quality tables
\usepackage{amsfonts}       % blackboard math symbols
\usepackage{nicefrac}       % compact symbols for 1/2, etc.
\usepackage{microtype}      % microtypography
\usepackage{xcolor}         % colors
\usepackage{amsmath,amssymb}
\usepackage{makecell}
\usepackage{graphicx}
\usepackage{tcolorbox}
\title{Grounding Long Financial Text with Semantically-Enriched Tables via Multimodal Learning}

\author{%
  Xiaoyu Hu\thanks{Use footnote for providing further information
    about author (webpage, alternative address)---\emph{not} for acknowledging
    funding agencies.} \\
  Department of Computer Science\\
  Cranberry-Lemon University\\
  Pittsburgh, PA 15213 \\
  \texttt{hippo@cs.cranberry-lemon.edu} \\
}

\begin{document}
\nolinenumbers

\maketitle

\begin{abstract}
 We propose a multimodal transformer framework for efficiently modeling long financial documents that combine structured tabular data with noisy, unstructured textual disclosures. The framework represents financial tables as semantically enriched token sequences and integrates them with long-form text through a cross-attention mechanism that explicitly captures cross-modal dependencies. In particular, we augment each numerical variable with a semantic embedding derived from its textual definition, enabling the model to encode both numerical magnitude and economic meaning. This design yields a more expressive and parameter-efficient representation of tabular data compared to value-only embeddings. Using a large-scale dataset of financial statements and MD\&A disclosures, we show that structured numerical features remain the dominant source of predictive signal; however, textual information provides consistent incremental gains when incorporated via cross-attention rather than simple concatenation. Our results further demonstrate that noisy textual disclosures are most useful when conditioned on a stable, semantically structured tabular representation. In particular, using frozen variable-level tabular representations as cross-attention queries enables the model to weight MD\&A chunks differently for each financial variable and observation. These findings suggest that cross-attention provides an efficient mechanism for table-conditioned aggregation of long-form narrative evidence.

\end{abstract}

\section{Introduction}

The rapid advancement of Large Language Models (LLMs) has sparked significant interest in their cross-domain applications. Recent research has explored LLMs within the financial sector for tasks such as sentiment analysis \cite{wang2023fingpt}, question answering \cite{zhao2024docmath}, and algorithmic trading \cite{yu2024fincon}. Despite these strides, many existing studies are oversimplified, relying on truncated contexts that fail to reflect real-world investment scenarios. Specifically, current financial prediction models often focus on fine-tuning generative architectures (e.g., GPT) on short-form unstructured data, such as tweets \cite{zhang2023instruct, xie2023pixiu, yang2023fingpt}. In contrast, professional investors typically analyze comprehensive documents like 10-K filings, which integrate structured numerical tables with extensive narrative disclosures. While some efforts have been made to leverage LLMs for stock performance forecasting, these approaches predominantly utilize generative models that ingest tabular and textual data as raw context. Such methods not only struggle with the complexity of multimodal integration but are also prone to look-ahead bias \citep{levy2024caution}, undermining their practical utility. To address these limitations, we propose a novel multimodal transformer framework to efficiently represent financial documents.

Traditionally, financial prediction task has relied heavily on structured data—the "hard" numbers found in balance sheets, income statements, and cash flow reports. However, the modern financial landscape is increasingly defined by "soft" information: the narrative disclosures, strategic commentaries, and qualitative nuances embedded in unstructured text. Despite the availability of both modalities, effectively integrating them into a unified, high-fidelity representation remains a significant challenge for existing machine learning architectures. Current research in financial forecasting often faces a "modality gap." Traditional econometrics and tree-based ensembles, such as XGBoost and LightGBM, excel at processing structured tabular data but are inherently incapable of ingesting narrative context. Conversely, the rise of Large Language Models (LLMs) and transformer-based encoders like BERT has revolutionized textual analysis, yet these models often struggle to maintain the rigid precision required for numerical tabular data. When these two worlds meet, the standard approach is often simple feature concatenation, which fails to account for the complex, non-linear interactions between a company’s reported figures and its qualitative disclosures. Furthermore, financial documents like the Management’s Discussion and Analysis (MD\&A) sections of 10-K filings—are characterized by a low signal-to-noise ratio. Disclosures are often lengthy, repetitive, and subject to "boilerplate" language that can obscure economically relevant information. Our findings suggest that unstructured financial text is most predictive not in isolation, but when it is grounded in the context of structured financial tables. Without a numerical anchor, the narrative is often too noisy to provide a stable signal for earnings prediction.

Specifically, our architecture comprises two primary components: a semantically-enriched tabular encoder and a hierarchical textual segment encoder. The tabular module innovates by projecting each financial variable into a high-dimensional space that fuses numerical magnitude with a semantic embedding of the variable's definition. A subsequent transformer layer learns the dependencies between variables across all tables within a document, creating a cohesive, context-aware representation of the firm's financial state. For the textual modality, we employ a hierarchical approach: long-form narrative disclosures, such as the Management’s Discussion and Analysis (MD\&A), are partitioned into fixed-length segments, encoded via ModernBERT, and pooled into segment-level representations. These two streams are integrated via a phased cross-attention mechanism, where the tabular embeddings serve as the query ($Q$) to selectively filter and aggregate informative signals from the textual keys and values ($K, V$). To evaluate this framework, we utilize a large-scale dataset of corporate 10-K filings, encompassing both structured financial statement variables and unstructured narrative text (MD\&A). The experimental results indicate that our framework outperform existing tabular multimodal deep learning architecture as well as tree-based models. The ablation study suggest only when the tabular encoder is trained first and frozen in cross-attention layer can unstructured texts contribute to final document representation. This architectural design allows the model to ground noisy, long-form text in the structural stability of financial tables, providing a more robust and expressive representation for downstream tasks like earnings prediction. We summarize the contributions of our paper as follows:

\begin{itemize}
    \item \textbf{Semantically-Enriched Tabular Representation}: We introduce an efficient tabular encoding strategy that maps financial variables into a unified document-level vector by fusing numerical magnitudes with their semantic definitions.
    \item \textbf{Grounding-Based Multimodal Framework}: We propose a novel framework that integrates structured tables and long-form unstructured text, utilizing a cross-attention mechanism to capture complex inter-modal dependencies for long financial documents.
    \item \textbf{Comprehensive Financial Benchmark}: We contribute a multimodal dataset comprising structured tables and extensive narrative disclosures, designed to simulate high-fidelity, real-world financial decision-making scenarios.
\end{itemize}

\section{Related Work}

\subsection{The Applications of LLMs in Finance Domain}

Textual data represents a cornerstone of unstructured information in the financial domain, prompting a robust body of literature dedicated to leveraging Large Language Models (LLMs) for financial analysis. While tasks like sentiment analysis—detecting positive, negative, or neutral tones to inform investment decisions—are well-established, existing benchmarks such as FinQA \citep{chen2022convfinqa} and FinancialPhraseBank \citep{malo2014good} face significant constraints. Current datasets are primarily characterized by short-context snippets, often constructed by concatenating isolated news headlines or tweets. Such formats fail to replicate the long-context environment of professional investing, where decisions are derived from extensive documentation rather than disjointed sentences. Furthermore, existing financial task benchmarks \citep{li2025investorbench} are frequently restricted to a handful of high-profile technology firms (e.g., Apple, Google), which precludes a comprehensive evaluation of LLM performance across diverse market sectors. Lastly, the majority of contemporary financial prediction frameworks \citep{xie2023pixiu, yang2023fingpt} utilize generative architectures (e.g., GPT) via in-context learning. By prepending a mixture of numerical and textual data into a single prompt for prediction, these models may inadvertently introduce look-ahead bias \citep{levy2024caution}, as temporal boundaries are often blurred within the context window.

\subsection{Tabular Transformer-Based Models}

Recent advances in representation learning for structured data have led to the development of transformer-based models tailored to tabular domains. Among these, the TabTransformer \cite{huang2020tabtransformer} extends the transformer architecture to tabular data by applying self-attention over feature embeddings, enabling the model to capture complex inter-feature dependencies. This approach contrasts with traditional machine learning methods such as gradient boosting and linear models, which often rely on handcrafted features or assume limited interaction structures. Subsequent studies \cite{arazi2025tabstar} have demonstrated that transformer-based tabular models can achieve competitive or superior performance, particularly in settings with heterogeneous feature types and rich semantic representation for numerical variables. Variants such as FT-Transformer \citep{gorishniy2021revisiting} further refine this framework by incorporating both numerical and categorical features into a unified attention-based architecture. In the context of financial prediction tasks, where interactions among accounting variables are often nonlinear and context-dependent, these models offer a principled way to learn feature representations directly from data. Overall, the emerging literature suggests that attention-based architectures provide a flexible and powerful alternative to conventional approaches for modeling tabular data.

\subsection{Multimodal Cross-Attention Mechanism}

A growing body of literature has explored multimodal transformer architectures as a principled framework for integrating heterogeneous data sources, particularly through the use of cross-attention mechanisms \citep{chen2021crossvit, tsai2019multimodal}. Prior work in vision-language tasks \citep{kim2021vilt}, such as visual question answering and image captioning, demonstrates that cross-attention layers can effectively learn fine-grained correspondences between modalities, significantly improving downstream performance. More recent studies extend these ideas to structured data settings, showing that cross-modal attention can uncover interactions between numerical and unstructured inputs, such as clinical records paired with physician notes or financial variables paired with narrative disclosures. Furthermore, hierarchical and multi-head cross-attention mechanisms enable the model to capture interactions at different levels of abstraction, while preserving modality-specific representations. Collectively, this literature highlights that the key advantage of multimodal transformers lies not merely in combining multiple data sources, but in explicitly modeling their interactions through attention-based alignment, offering a flexible and scalable framework for complex prediction tasks involving heterogeneous information. However, prior literature's multi-modal setting is limited in short contexts (short sentiment + numbers), in contrast, we focus on multimodal long-context document (long tables + long narrative texts) which is more common in finance.

\section{Model Design}

\subsection{Encoding Structured Financial Tables}

Unlike prior literature that often treat numeric features as themselves, we propose a new model architecture to encode tabular financial data by integrating natural language semantics with numerical values. The model processes each table as a sequence of row-level embeddings, where each row $i$ is defined by a text-based feature name $T_i$ and a corresponding scalar value $v_i$. For each row, the semantic context of the feature name is captured using a light pre-trained sentence transformer encoder (all-MiniLM-L6-v2), which produces a fixed-length vector $\mathbf{e}^{text}_i \in \mathbb{R}^{384}$. This vector is subsequently refined through a linear projection and layer normalization to match the model's internal dimension $d_{model}$:

\begin{equation}
\mathbf{z}^{text}_i = \text{LayerNorm}(\mathbf{W}_r \cdot \mathbf{e}^{text}_i + \mathbf{b}_r)
\end{equation}

Simultaneously, the scalar value $v_i$ is mapped into a high-dimensional space with the same dimension as $z^{text}_i$ via a linear Encoder. This module utilizes a multi-layer perceptron (MLP) with GELU and Tanh activations to capture the magnitude and scale of the financial metric:

\begin{equation}
    \mathbf{z}_{i}^{num} = \text{Tanh}(\text{MLP}_{num}(v_i))
\end{equation}

The final "cell embedding" $\mathbf{x}_i$ for a given row is then generated by the element-wise addition of the text and numerical representations, followed by a final normalization step to ensure stable gradients during training:

\begin{equation}
\mathbf{x}_i = \text{LayerNorm}(\mathbf{z}^{\text{text}}_i + \mathbf{z}^{\text{num}}_i)
\end{equation}

To capture the inter-dependencies between various financial line items (e.g., the relationship between \textit{Revenue} and \textit{Cost of Goods Sold}), the table is represented as a sequence $\mathbf{X} = [\mathbf{x}_{cls}, \mathbf{x}_1, \mathbf{x}_2, \dots, \mathbf{x}_n]$, where $\mathbf{x}_{cls}$ is a learnable table summarization token prepended to the sequence. This sequence is processed through a $L$-layer Transformer Encoder. Each layer performs multi-head self-attention ($\text{MHSA}$) and point-wise feed-forward ($\text{FFN}$) operations:

\begin{equation}
\tilde{\mathbf{H}}^{(l)} = \mathrm{MHSA}(\mathbf{H}^{(l-1)}) + \mathbf{H}^{(l-1)}
\end{equation}

\begin{equation}
\mathbf{H}^{(l)} = \mathrm{FFN}(\mathrm{LayerNorm}(\tilde{\mathbf{H}}^{(l)})) + \tilde{\mathbf{H}}^{(l)}
\end{equation}

The global context of the table is aggregated in the final state of the \texttt{[CLS]} token, $\mathbf{h}^{(L)}_{cls}$. This vector can represent the whole table and serve as the input to our downstream task --- earnings direction prediction. We can either concat $\mathbf{h}^{(L)}_{cls}$ with other features or directly add a linear classification head to produce the final prediction $y$:

\begin{equation}
y = \text{Softmax}(\mathbf{W}_h \mathbf{h}^{(L)}_{\text{cls}} + \mathbf{b}_h)
\end{equation}

\subsection{Encoding Unstructured Financial Texts}

Textual representation learning has undergone substantial evolution, progressing from simple frequency-based methods to context-aware transformer architectures. Early approaches to encoding textual information primarily rely on word occurrence statistics and treat words as independent units. While these methods offer computational simplicity and interpretability, they are inherently limited in their ability to capture semantic meaning, contextual dependencies, and higher-order linguistic structures. The introduction of transformer-based language models, such as BERT, represents a significant advancement by enabling context-sensitive encoding through self-attention mechanisms. To encode long-form financial narratives without document-level truncation, we adopt a hierarchical text-encoding strategy. Specifically, each MD\&A document \(D\) is segmented into a sequence of non-overlapping chunks \(\{c_1,c_2,\ldots,c_{n_D}\}\), where \(n_D\) denotes the number of chunks produced for document \(D\). Each chunk contains at most \(2{,}048\) tokens. We preserve paragraph boundaries whenever a complete paragraph fits within this limit; if a paragraph exceeds \(2{,}048\) tokens, it is divided into two or more chunks at sentence boundaries so that no sentence is split across chunks. Each chunk \(c_i\) is independently processed by the ModernBERT encoder to obtain a chunk-level representation \(\mathbf{h}_i\). We impose no global upper bound on \(n_D\) and therefore retain all chunks generated from each document. Within each mini-batch \(\mathcal{B}\), the chunk dimension is dynamically set to \(N_{\mathcal{B}}=\max_{D\in\mathcal{B}}n_D\); documents with fewer chunks are padded to this length, and padded positions are excluded using an attention mask. For the text-only model, the document-level representation is computed by mean-pooling over only the valid chunk representations, \(\mathbf{z}=\frac{1}{n_D}\sum_{i=1}^{n_D}\mathbf{h}_i\). Finally, a linear classification head maps \(\mathbf{z}\) to two logits, from which the predicted earnings-direction label \(\hat{y}\) is obtained.

\subsection{Encoding whole financial document}

To represent long financial documents that integrate structured financial tables with unstructured narrative disclosures, we employ cross-attention to model context-dependent interactions between the two modalities. The architecture contains a text branch, in which the MD\&A is encoded as a sequence of chunk representations using ModernBERT, and a tabular branch, in which financial variables are encoded as a sequence of semantically enriched tabular tokens. Unlike late-fusion approaches that first compress each modality into a single vector, our model retains both sequences during cross-modal interaction. Let $
H_{\mathrm{tab}}
=
[h^{\mathrm{tab}}_1,\ldots,h^{\mathrm{tab}}_m]
\in\mathbb{R}^{m\times d}
$ denote the sequence of contextualized tabular tokens produced by the tabular Transformer, where \(m\) is the number of financial variables and each token represents one financial row by combining its numerical value with the semantic representation of its variable definition. Similarly, let $
H_{\mathrm{text}}
=
[h^{\mathrm{text}}_1,\ldots,h^{\mathrm{text}}_n]
\in\mathbb{R}^{n\times d}
$ denote the sequence of \(n\) MD\&A chunk representations produced by the hierarchical text encoder. We investigate two cross-attention directions, distinguished by which modality provides the queries.

In the tabular-query configuration, every financial-variable token acts as an independent query, while the MD\&A chunk representations provide the keys and values:

\begin{equation}
\label{tab-query}
Q_{\mathrm{tab}}=H_{\mathrm{tab}}W_Q^{\mathrm{tab}},
\qquad
K_{\mathrm{text}}=H_{\mathrm{text}}W_K^{\mathrm{text}},
\qquad
V_{\mathrm{text}}=H_{\mathrm{text}}W_V^{\mathrm{text}}.
\end{equation}

The corresponding attention matrix and output sequence are

\begin{equation}
\label{tab-query-attention}
A_{\mathrm{tab}\leftarrow\mathrm{text}}=
\operatorname{softmax}
\left(
\frac{
Q_{\mathrm{tab}}K_{\mathrm{text}}^\top
}{
\sqrt{d_k}
}
\right)
\in\mathbb{R}^{m\times n},
\end{equation}

\begin{equation}
Z_{\mathrm{tab}\leftarrow\mathrm{text}} = 
A_{\mathrm{tab}\leftarrow\mathrm{text}}V_{\mathrm{text}}
\in\mathbb{R}^{m\times d_v}.
\end{equation}

Each element
\([A_{\mathrm{tab}\leftarrow\mathrm{text}}]_{ij}\)
represents the attention weight assigned by financial-variable token \(i\) to MD\&A chunk \(j\). This configuration grounds the resulting multimodal representation in structured tabular information while selectively incorporating relevant evidence from the narrative disclosures.

In the text-query configuration, every MD\&A chunk representation acts as an independent query, while the tabular tokens provide the keys and values:

\begin{equation}
\label{tab-query}
Q_{\mathrm{text}}=H_{\mathrm{text}}W_Q^{\mathrm{text}},
\qquad
K_{\mathrm{tab}}=H_{\mathrm{tab}}W_K^{\mathrm{tab}},
\qquad
V_{\mathrm{tab}}=H_{\mathrm{tab}}W_V^{\mathrm{tab}}.
\end{equation}

The corresponding attention matrix and output sequence are

\begin{equation}
\label{text-query-attention}
A_{\mathrm{text}\leftarrow\mathrm{tab}}=
\operatorname{softmax}
\left(
\frac{
Q_{\mathrm{text}}K_{\mathrm{tab}}^\top
}{
\sqrt{d_k}
}
\right)
\in\mathbb{R}^{n\times m},
\end{equation}

\begin{equation}
Z_{\mathrm{text}\leftarrow\mathrm{tab}}=
A_{\mathrm{text}\leftarrow\mathrm{tab}}V_{\mathrm{tab}}
\in\mathbb{R}^{n\times d_v}.
\end{equation}

Here,
\([A_{\mathrm{text}\leftarrow\mathrm{tab}}]_{ji}\)
represents the importance assigned by MD\&A chunk \(j\) to financial-variable token \(i\). This configuration therefore produces a sequence of table-conditioned textual chunk representations, with each chunk receiving its own weighted aggregation of the financial variables. The two configurations model complementary directions of cross-modal interaction. Tabular-query attention identifies which MD\&A chunks are relevant to each financial variable, whereas text-query attention identifies which financial variables are relevant to each MD\&A chunk. Both configurations therefore operate at the financial-variable and text-chunk levels, providing more structured cross-modal interaction than the direct concatenation of pooled document-level representations. For both cross-attention configurations, we follow the standard multi-head Transformer architecture. The outputs from the \(H\) attention heads are concatenated and projected to the shared hidden
dimension:$
\bar{Z}
=
\operatorname{Concat}
\left(
Z^{(1)},\ldots,Z^{(H)}
\right)W_O.
$
Let \(H_Q\) denote the original query sequence and \(L\) its length. For table-to-text attention, \(H_Q=H_{\mathrm{tab}}\) and \(L=m\),
whereas for text-to-table attention, \(H_Q=H_{\mathrm{text}}\) and
\(L=n\). The projected attention output is integrated with the query
sequence through a residual connection and layer normalization, followed by a position-wise feed-forward network with a second residual connection and layer normalization.
\begin{equation}
\widetilde{H}
=
\operatorname{LayerNorm}
\left(
H_Q+\operatorname{Dropout}(\bar{Z})
\right),
\end{equation}
\begin{equation}
H_{\mathrm{out}}
=
\operatorname{LayerNorm}
\left(
\widetilde{H}
+
\operatorname{Dropout}
\left(
\operatorname{FFN}(\widetilde{H})
\right)
\right).
\end{equation}

Consequently, the table-to-text configuration produces
\(H_{\mathrm{out}}\in\mathbb{R}^{m\times d}\), while the
text-to-table configuration produces
\(H_{\mathrm{out}}\in\mathbb{R}^{n\times d}\). To aggregate the variable-length cross-modal output, we prepend a learned fusion token
\(\mathbf{c}_{\mathrm{fuse}}\in\mathbb{R}^{d}\) to
\(H_{\mathrm{out}}\) and process the augmented sequence using a Transformer encoder layer:
\begin{equation}
G
=
\operatorname{TransformerEncoder}
\left(
[\mathbf{c}_{\mathrm{fuse}};H_{\mathrm{out}}]
\right),
\qquad
\mathbf{z}_{\mathrm{fused}}=G_0,
\end{equation}
where \(G_0\) is the contextualized representation of the fusion
token. An attention mask prevents the fusion token from attending to
padded positions. In the table-to-text configuration, the fusion
token aggregates information from the \(m\) text-conditioned
financial-variable tokens. In the text-to-table configuration, it
aggregates information from the \(n\) table-conditioned MD\&A chunks.
Finally, a linear classification head maps the resulting representation
to two logits:
\begin{equation}
\boldsymbol{\ell}
=
W_{\mathrm{cls}}\mathbf{z}_{\mathrm{fused}}
+
\mathbf{b}_{\mathrm{cls}}
\in\mathbb{R}^{2},
\qquad
\mathbf{p}
=
\operatorname{softmax}(\boldsymbol{\ell}).
\end{equation}

\section{Experiments Setup}

\paragraph{Datasets} Our dataset consists of 50,849 firm-year observations constructed from corporate 10-K filings and corresponding financial statements covering fiscal years 2000--2025. Each observation for firm \(i\) and predictor fiscal year \(t\) contains only the MD\&A narrative and structured financial variables associated with year \(t\). We exclude MD\&A sections containing fewer than 512 tokens; the retained sections contain approximately 15,000 tokens on average, motivating the use of a long-document architecture. We measure earnings using annual net income (Compustat item \(NI\)) and define the next-year earnings-direction label as \(y_{i,t}=\mathbb{I}(NI_{i,t+1}-NI_{i,t}>0)\), where \(y_{i,t}=1\) indicates an increase in net income and \(y_{i,t}=0\) indicates no change or a decrease. Observations without valid net-income values in either year are excluded. We partition the data according to predictor fiscal year, using 2000--2017 for training, 2018--2019 for validation, and 2020--2024 for held-out testing. For observations from fiscal year 2024, the 2025 net-income value is used only to construct the target label; no 2025 MD\&A text or financial variables are included as predictors. Appendix~\ref{appendix_data} provides a detailed sample observation, while Tables~\ref{tab:Sample} and~\ref{tab:SampleLabel} report the industry and label distributions, respectively. The sample covers all Global Industry Classification Standard sectors, and the training and test sets have similar class distributions, with approximately 55\% positive and 45\% negative observations.

\paragraph{Benchmarks} We compare our model architecture with previous tabular transformer model including Resnet \citep{rubachev2022revisiting}, FT-transformer \citep{rubachev2022revisiting}, Tabnet \citep{arik2021tabnet}, TabSTAR \citep{arazi2025tabstar} and Tabtransformer \citep{huang2020tabtransformer}, Trompt \citep{chen2023trompt}. These models are specifically designed to ingest multimodal tabular data containing both numerical and textual features. These models encode numeric columns and the text column into a shared feature space, then jointly learn from them inside the backbone model; the text is usually represented as a document-level feature as similar structural level as every numeric column. To ensure a fair multimodal comparison, every tabular baseline is provided with the same representation of the complete MD\&A document using the hierarchical ModernBERT procedure described in Section~3.2. This document vector is then concatenated with the firm’s numerical financial features and provided to each tabular model, including ResNet, FT-Transformer, and TabTransformer etc.  Furthermore, we benchmark against industry-standard tree-based ensembles: Random Forest, XGBoost \citep{chen2016xgboost}, LightGBM \citep{ke2017lightgbm}, and CatBoost \citep{prokhorenkova2018catboost}. Although tree-based methods frequently achieve competitive performance on tabular tasks, they lack the native ability to process multimodal inputs or generate cohesive latent representations of a table. For tree-based models, we only concatenate the frozen text embeddings as features. To isolate the impact of specific design choices, we ablate several model features: specifically, we assess the performance differential between static and trainable text embeddings, the utility of auxiliary text columns, and the efficacy of complex cross-modal interactions over basic feature concatenation.

\paragraph{Model Training and Evaluation}We employ systematic hyperparameter tuning using validation set to optimize model performance across all experimental settings. For tree-based models, we use grid search to get optimal hyperparameters and report the best performance for each model. For deep learning models, we consider hyperparameters including batch sizes, dropout rates, learning rates etc. The detailed hyperparameter search is in appendix \ref{appendix_B}. The optimal set of hyperparameters is selected based on the accuracy achieved on the validation set. For all deep learning model training experiments, we adopt the Adam optimizer with weight decay ($\beta=0.9$, $\lambda = 0.01$). We utilize a parameter-efficient fine-tuning (PEFT) strategy, specifically Low-Rank Adaptation (LoRA), to finetune the text encoder (ModernBERT) in all model systems. This allows for effective domain specialization without the substantial computational overhead. We implemented early stopping with a patience of 2 epochs, monitoring validation loss to prevent overfitting, with a maximum limit of 20 epochs. Since it is a binary classification task, we use ROC-AUC, Accuracy and macro-F1 score to evaluate model performance. We run each model using five random seeds and conduct ten runs for each seed, yielding 50 results in total. For each performance metric, we calculate the mean and sample standard deviation across all 50 runs. The training was utilizing 4 NVIDIA A100 80GB GPUs.

\paragraph{Experiment Results} Table \ref{tab:earnings_new} summarizes the performance of various tabular deep learning architectures and tree-based models on the earnings prediction task. Our results demonstrate that current tabular deep learning frameworks consistently achieve lower Accuracy, AUC, and F1 scores compared to our proposed architecture when processing integrated numerical data and unstructured MD\&A text. Specifically, models such as TabTransformer and FT-Transformer yield approximately 60\% accuracy, underperforming our framework by 6 percentage points. Interestingly, tree-based models tend to outperform tabular deep learning models in both accuracy and F1 score due to the fact that feature selection is built in tree-based models. However, our approach outperforms all benchmarked tree-based models, including Random Forest, LightGBM, and XGBoost. These improvements stem from two primary architectural innovations: first, our framework incorporates numerical variable semantics into the table representation, whereas baseline models typically rely on raw numerical inputs and simple imputation methods for missing values. Second, recognizing that financial text often suffers from a low signal-to-noise ratio, we employ a cross-attention mechanism rather than simple feature concatenation. This enables a more dynamic and weighted integration of tabular and textual modalities within the holistic document representation.

Table \ref{tab:Texttable} presents the results of our ablation study evaluating various architectural components. In unimodal configurations, the financial tabular embeddings significantly outperform text-based representations using either attention or mean pooling. Specifically, the tabular-only model achieves an accuracy of 65.52\% and an AUC of 70.05\%, whereas the text-only baseline yields only 57.32\% and 59.29\%, respectively. These results suggest that structured financial data provides a more robust signal for this task than unstructured textual descriptions. In multimodal settings, the integration of textual and tabular data presents significant complexity. First, simple concatenation of tabular and textual embeddings—regardless of whether the encoders are frozen or fine-tuned—consistently underperforms the unimodal tabular model. For example, joint training via concatenation yields an accuracy of 63.57\%, approximately 2 percentage points lower than the tabular-only baseline (65.52\%). Second, within the cross-attention framework, using textual embeddings as the query ($Q$) vector consistently results in degraded performance compared to using tabular queries, irrespective of the learning regime. Superior performance is achieved only by employing a two-stage approach: pre-training tabular embeddings, freezing their weights, and subsequently utilizing them as the query ($Q$) in the cross-attention mechanism. This configuration achieves our peak performance of 66.64\% accuracy and 73.04\% AUC. These results suggest that while textual disclosures provide complementary signals, they are inherently noisier than structured financial data and are most effective when integrated through a schema that prioritizes a stable tabular foundation.

\begin{table}[htbp]
\centering
\setlength{\tabcolsep}{15pt}
\small

%\caption{Model Herding with 'fake' rating}
\begin{tabular}{lccc}
%\begin{tabularx}{\textwidth}{l *{10}{s}}
\toprule
Model  &
AUC &
Acc &
Macro-F1 \\
\hline
\multicolumn{4}{c}{\textit{Tabular Deep Learning Model}} \\
\hline
Resnet \citep{rubachev2022revisiting} &$0.6395 \pm 0.006$ & $0.5784 \pm 0.009$ & $0.5382 \pm 0.010$\\
FT-Transformer \citep{rubachev2022revisiting} & $0.7025 \pm 0.004$ & $0.6005 \pm 0.006$ &$0.5702 \pm 0.009$ \\
TabTransformer \citep{huang2020tabtransformer}& $0.6546 \pm 0.006$ & $0.6055 \pm 0.004$& $0.5520 \pm 0.009$ \\
TabNet \citep{arik2021tabnet} &$0.6388 \pm 0.006$ & $0.6033 \pm 0.005$& $0.5473 \pm 0.007$\\
TabSTAR \citep{arazi2025tabstar} & $0.6929 \pm 0.005$& $0.6453 \pm 0.004$& $0.6024 \pm 0.005$\\
Trompt \citep{chen2023trompt} &$0.6838 \pm 0.008$ & $0.6192 \pm 0.003$ & $0.5835 \pm 0.006$\\
\textbf{Ours} & \textbf{$0.7304 \pm 0.003$}  & \textbf{$0.6664 \pm 0.002$} & \textbf{$0.6250 \pm 0.004$} \\
\hline
\multicolumn{4}{c}{\textit{Tree-Based Models}} \\
\hline 
Random Forest &$0.6828 \pm 0.006$ &$0.6344 \pm 0.005$ &$0.6012 \pm 0.006$ \\
CatBoost \citep{prokhorenkova2018catboost}&$0.6837 \pm 0.004$&$0.6359 \pm 0.003$ &$0.6034 \pm 0.004$ \\
LightGBM \citep{ke2017lightgbm}&$0.6899 \pm 0.004$& $0.6442 \pm 0.003$ &$0.6119 \pm 0.006$ \\
Xgboost \citep{chen2016xgboost}& $0.6958 \pm 0.005$& $0.6453 \pm 0.004$& $0.6121 \pm 0.007$ \\

\bottomrule
\end{tabular}
\caption{Comparison of model performance between benchmark models including tabular deep learning models and tree-based models in multi-modal setting (tabular + text) under earnings prediction task. Results are reported as mean \(\pm\) the half-width of the 95\% standard normal confidence interval over 50 runs.}
\label{tab:earnings_new}
\end{table}

\begin{table}[htbp]
\centering
\small
\begin{tabular}{p{8cm} ccc}
\toprule
\textbf{Model Design}  & \textbf{ROC-AUC} & \textbf{Acc} & \textbf{Macro-F1} \\
\midrule

% First row: Citation aligned with the first (comparison) MLP
        \makecell{Learned Tabular Embedding} & \textbf{$0.6984 \pm 0.002$} & \textbf{$0.6549 \pm 0.001$} & \textbf{$0.6032 \pm 0.002$} \\ [12pt] 

        \makecell{Learned Text Embedding + attention pooling} & $0.5903 \pm 0.002$ & $0.5748 \pm 0.003$  & $0.5365 \pm 0.003$ \\ [12pt] 

        \makecell{Learned Text Embedding + mean pooling} & $0.5807 \pm 0.003$ & $0.5700 \pm 0.004$ & $0.5263 \pm 0.002$ \\ [12pt]
        
         \makecell{Freeze Text Embedding + \\ Learned Tabular Embedding + Concatenation } & $0.6765 \pm 0.002$   & $0.6483 \pm 0.001$  & $0.5987 \pm 0.002$ \\ [12pt] \\

        \makecell{Learned Text Embedding + \\ Freeze Learned Tabular Embedding + Concatenation } & $0.5892 \pm 0.002$ & $0.6442 \pm 0.004$ & $0.5515 \pm 0.002$ \\ [12pt] \\
        
         \makecell{Learned Text Embedding + \\ Learned Tabular Embedding +  Concatenation} & $0.6837 \pm 0.002$& $0.6325 \pm 0.003$ & $0.6028 \pm 0.003$ \\ [12pt] \\

% Third row: Tabular Transformer
         \makecell{Learned Text Embedding (Q) + \\Learned Tabular Embedding (K, V) +Cross-Attention} & $0.6766 \pm 0.003$ & $0.6248 \pm 0.002$ & $0.5956 \pm 0.002$ \\ [12pt] \\

         \makecell{Freeze Text Embedding (Q) + \\ Learned Tabular Embedding (K, V) +Cross-Attention} & $0.6634 \pm 0.002$ & $0.6051 \pm 0.002$ & $0.5870 \pm 0.002$\\ [12pt] \\ 

         \makecell{Learned Tabular Embedding (Q) + \\Learned Text Embedding (K, V) + Cross-Attention} & $0.7073 \pm 0.003$ & $0.6533 \pm 0.002$ & $0.6094 \pm 0.002$ \\ [12pt] \\

         \makecell{Freeze Learned Tabular Embedding (Q) + \\ Learned Text Embedding (K, V) + Cross-Attention} & \textbf{$0.7304 \pm 0.003$} & \textbf{$0.6664 \pm 0.002$} & \textbf{$0.6250 \pm 0.004$}\\ [12pt] 

        \bottomrule
    \end{tabular}

    \caption{Ablation study for different model components. This table compares different model architecture choices (freeze text encoder vs unfreeze text encoder; concatenation vs cross-attention; freeze tabular transformer vs unfreeze tabular transformer) in earnings prediction performance. Results are reported as mean \(\pm\) the half-width of the 95\% standard normal confidence interval over 50 runs.
} 
\label{tab:Texttable}    

\end{table}

\section{Conclusion}

Our study provides a comprehensive examination of the roles played by structured financial variables and unstructured textual disclosures in earnings prediction, with a specific focus on multimodal integration. We first propose a tabular model that can efficiently represent the whole structured tables. Notably, we observe that due to the low signal-to-noise ratio inherent in financial text, unstructured disclosures are most effective when grounded in the context of structured financial tables. To achieve this, we implement a phased training strategy: we first optimize the tabular module, then freeze it to serve as a query ($Q$) vector. This allows the model to selectively and dynamically attend to textual features ($K, V$) via a cross-attention mechanism, resulting in a more flexible and expressive representation of long-form financial documents. However, our study only provides one financial task (earnings prediction) using large multimodal dataset, shedding light on future research studying more financial tasks based on multimodal data like risk management, stock prediction.

\newpage
\bibliographystyle{plain}
\bibliography{references}

%%%%%%%%%%%%%%%%%%%%%%%%%%%%%%%%%%%%%%%%%%%%%%%%%%%%%%%%%%%%

\appendix

\section{Dataset Example}\label{appendix_data}

\subsection{Management Discussion and Management (MD\&A)}
%%%%%%%%%%%%%%%%%%%%%%%%%%%%%%%%%%%%%%%%%%%%%%%%%%%%%%%%%%%%
The Management’s Discussion and Analysis (MD\&A) section is a mandatory, narrative component of a company’s annual report (Form 10-K) or quarterly report (Form 10-Q), providing investors with a qualitative look into the financial and operational performance from the perspective of the management team. Unlike the audited financial statements, which provide a quantitative snapshot of the past, the MD\&A serves to "bridge the gap" between raw data and business reality by explaining the "why" behind the numbers. It details the company's liquidity, capital resources, and results of operations, while specifically highlighting known trends, commitments, or uncertainties—such as inflation, supply chain disruptions, or regulatory changes—that are reasonably likely to impact future performance. By offering this critical context, the MD\&A helps stakeholders assess the quality of earnings and the sustainability of the company's competitive position. Prior literature indicates when a company’s financial fundamentals (like earnings) are poor, managers often "pad" the MD\&A with excessive positive sentiment to mislead investors or stabilize stock prices, leading to biased narriative disclosure \citep{huang2014tone}. They found that firms with poor performance or high uncertainty often use high levels of "boilerplate" text (copy-pasting from the previous year). This allows them to comply with the letter of the law while avoiding the disclosure of specific, evolving risks \citep{brown2011large}. Figure \ref{fig:apple-report} gives an example MD\&A section from Apple Inc. 10K filling. 

\begin{figure*}
\centering
\begin{tcolorbox}[colback=teal!5, colframe=teal, title= Context]
    
\textbf{Product, Service and Software Announcements}
The Company announces new product, service and software offerings at various times during the year. Significant announcements during fiscal year 2025
included the following: ... \\

\textbf{Fiscal Period}
The Company’s fiscal year is the 52- or 53-week period that ends on the last Saturday of September. An additional week is included in the first fiscal quarter
every five or six years to realign the Company’s fiscal quarters with calendar quarters, which occurred in the first quarter of 2023. The Company’s fiscal years
2025 and 2024 spanned 52 weeks each, whereas fiscal year 2023 spanned 53 weeks. \\

\textbf{Macroeconomic Conditions} \\
Macroeconomic conditions, including inflation, interest rates and currency fluctuations, have directly and indirectly impacted, and could in the future materially
impact, the Company’s results of operations and financial condition.\\

\textbf{Tariffs and Other Measures} \\
Beginning in the second quarter of 2025, new U.S. Tariffs were announced, including additional tariffs on imports from China, India, Japan, South Korea, Taiwan,
Vietnam and the EU, among others. In response, several countries have imposed, or threatened to impose, reciprocal tariffs on imports from the U.S. and other
retaliatory measures. Various modifications to the U.S. Tariffs have been announced and further changes could be made in the future, which may include
additional sector-based tariffs or other measures. For example, the U.S. Department of Commerce has initiated an investigation under Section 232 of the Trade
Expansion Act of 1962, as amended, into, among other things, imports of semiconductors, semiconductor manufacturing equipment, and their derivative
products, including downstream products that contain semiconductors. Tariffs and other measures that are applied to the Company’s products or their
components can have a material adverse impact on the Company’s business, results of operations and financial condition, including impacting the Company’s
supply chain, the availability of rare earths and other raw materials and components, pricing and gross margin. The ultimate impact remains uncertain and will
depend on several factors, including whether additional or incremental U.S. Tariffs or other measures are announced or imposed, to what extent other countries
implement tariffs or other retaliatory measures in response, and the overall magnitude and duration of these measures. Trade and other international disputes
can have an adverse impact on the overall macroeconomic environment and result in shifts and reductions in consumer spending and negative consumer
sentiment for the Company’s products and services, all of which can further adversely affect the Company’s business and results of operations.\\

\textbf{Products and Services Performance}\\
iPhone \\
iPhone net sales increased during 2025 compared to 2024 due to higher net sales of Pro models.\\
Mac \\
Mac net sales increased during 2025 compared to 2024 primarily due to higher net sales of laptops and desktops.\\
iPad \\
iPad net sales increased during 2025 compared to 2024 primarily due to higher net sales of iPad Air, iPad mini and iPad, partially offset by lower net sales of iPad Pro. \\
Wearables, Home and Accessories \\
Wearables, Home and Accessories net sales decreased during 2025 compared to 2024 primarily due to lower net sales of Accessories and Wearables.\\
...
\end{tcolorbox}
\caption{A sample MD\&A from Apple Inc. 10K}
\label{fig:apple-report}
\end{figure*}

\subsection{Tabular Data}

\begin{figure}[htb]
\center{\includegraphics[width=15cm]{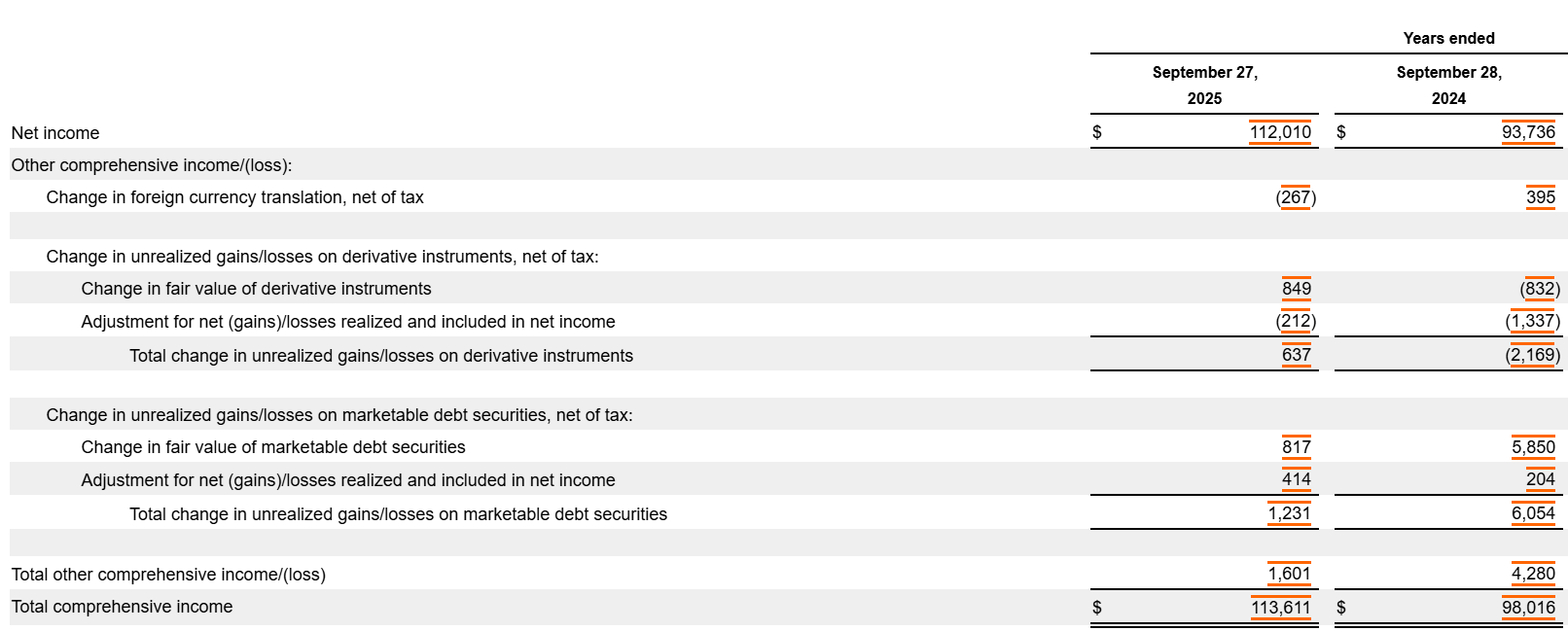}}   
\caption{\label{income_statement} Structured Table From Apple Inc. 10K}   
\end{figure}

Structured financial tabular data, mostly found in  the Financial Statements and Supplementary Data section, typically designated as Item 8 in a Form 10-K, constitutes the quantitative core of the annual report. This section presents the formal record of a company's financial activities and contains the four primary financial statements: the consolidated balance sheets, the statements of income (or operations), the statements of cash flows, and the statements of stockholders' equity. Crucially, these statements are prepared in accordance with Generally Accepted Accounting Principles (GAAP) and are accompanied by the Report of Independent Registered Public Accounting Firm, which provides the auditor's opinion on whether the statements fairly represent the company’s financial position. Beyond the raw numbers, this section includes extensive Notes to Consolidated Financial Statements. These footnotes are essential for transparency, as they disclose the specific accounting policies used, provide granular breakdowns of line items (such as debt maturity schedules or segment performance), and reveal potential liabilities or "off-balance-sheet" arrangements that are not immediately visible on the face of the statements. The Compustat database provides all financial statement items and detail variable definitions. Figure \ref{income_statement} gives an income statement example from Apple Inc. 10K.

\begin{table}[t]
    \centering
    % --- FIRST TABLE ---
    \begin{minipage}[t]{0.48\textwidth}
        \centering
        \caption{Industry Distribution} % This will be Table 1
        \label{tab:Sample}
        \begin{tabular}{lcc}
            \toprule
            Sector & Num & \% \\
            \midrule
            Technology & 9962 & 19.6\% \\
            Industrials & 8779 & 17.3\% \\
            Financials & 1744 & 3.4\% \\
            Healthcare & 10497 & 20.6\% \\
            Consumer Discretionary & 7238& 14.2\%\\
            Consumer Staples & 2405 & 4.7\% \\
            Utilities & 1293& 2.5\%\\
            Communication Services & 1900& 3.7\%\\
            Materials & 2620 & 5.2\% \\
            Energy & 3501& 6.9\%\\
            Real Estate & 910 & 1.8\%\\
            Total & 50849 & 100\% \\
            \bottomrule
        \end{tabular}
    \end{minipage}
    \hfill % Pushes the tables to the edges
    % --- SECOND TABLE ---
    \begin{minipage}[t]{0.48\textwidth}
        \centering
        \caption{Label Distribution} % This will be Table 2
        \label{tab:SampleLabel}
        \begin{tabular}{lcc}
            \toprule
            Label & Num & \% \\
            \midrule
            \textbf{Train+Validation Sample} \\
            Increase & 21471 & 55.47\% \\
            Decrease & 17236 & 44.53\% \\
            \midrule
            \textbf{Test Sample} \\
            Increase & 6732 & 55.44\% \\
            Decrease & 5410 & 44.56\% \\
            \bottomrule
        \end{tabular}
    \end{minipage}
\end{table}

\section{Hyperparameter Tuning}\label{appendix_B}

\subsection{Tabuar Transformer}

Since we use SentenceTransformer('all-MiniLM-L6-v2') to encode the semantic part, we align our model architecture with SentenceTransformer('all-MiniLM-L6-v2') which has d\_model (384), number of layers (6). We also tried using 192 d\_model, 768 d\_model by adding a linear projection layer to project 384 to lower/higher dimensions as well as 3 and 12 layers. The hyperparameter search space is shown in table \ref{tab:tabtransformer_hyperparameters}.

\begin{table}[htbp]
\centering
\caption{Tabular Transformer Model Search Space}
\label{tab:tabtransformer_hyperparameters}
\begin{tabular}{ll}
\hline
\textbf{Hyperparameter} & \textbf{Search Space} \\ \hline
$d\_model$ & $\mathcal{U}(192, 384, 768)$ \\
$Num\_Layers$ & $\mathcal{U}(3, 6, 12)$ \\
$Num\_Attention\_Layers$ & $ \mathcal{U}(8, 16, 32)$ \\
$Batch\_Size$ & $\mathcal{U}(16, 32, 64)$ \\
$Dropout\_out$ & $\mathcal{U}(0, 0.1, 0.2)$ \\
$Learning\_Rate$ & $\mathcal{U}($1e-5$, $2e-5$, $1e-4$)$ \\ \hline
\end{tabular}
\end{table}

\subsection{CatBoost}

We run CatBoost using \textit{catboost} package. We follow prior literature \cite{rubachev2022revisiting, arazi2025tabstar} setting  $early\_stopping\_rounds = 50, od\_pval = 0.001, iterations = 2000$ and do hyperparameter search \citep{hollmann2025accurate, arazi2025tabstar}. The hyperparameter search space is shown in table \ref{tab:catboost_hyperparameters}.

\begin{table}[htbp]
\centering
\caption{CatBoost-Tuned Hyperparameters Search Space}
\label{tab:catboost_hyperparameters}
\begin{tabular}{ll}
\hline
\textbf{Hyperparameter} & \textbf{Search Space} \\ \hline
learning\_rate & $\log \mathcal{U}(e^{-5}, 1)$ \\
random\_strength & $\mathcal{U}(\{1, 2, \dots, 20\})$ \\
l2\_leaf\_reg & $\log \mathcal{U}(1, 10)$ \\
bagging\_temperature & $\mathcal{U}(0.0, 1.0)$ \\
leaf\_estimation\_iterations & $\mathcal{U}(\{1, 2, \dots, 20\})$ \\
iterations & $\mathcal{U}(\{100, 101, \dots, 4000\})$ \\ \hline
\end{tabular}
\end{table}

\begin{table}[htbp]
\centering
\caption{XGBoost-Tuned Hyperparameters Search Space}
\label{tab:xgboost_hyperparameters}
\begin{tabular}{ll}
\hline
\textbf{Hyperparameter} & \textbf{Search Space} \\ \hline
learning\_rate & $\log \mathcal{U}(e^{-7}, 1)$ \\
max\_depth & $\mathcal{U}(\{1, 2, \dots, 10\})$ \\
subsample & $\mathcal{U}(0.2, 1)$ \\
colsample\_bytree & $\mathcal{U}(0.2, 1)$ \\
colsample\_bylevel & $\mathcal{U}(0.2, 1)$ \\
min\_child\_weight & $\log \mathcal{U}(e^{-16}, e^5)$ \\
alpha & $\log \mathcal{U}(e^{-16}, e^2)$ \\
reg\_lambda & $\log \mathcal{U}(e^{-16}, e^2)$ \\
gamma & $\log \mathcal{U}(e^{-16}, e^2)$ \\
n\_estimators & $\mathcal{U}(\{100, 101, \dots, 4000\})$ \\ \hline
\end{tabular}
\end{table}

\begin{table}[htbp]
\centering
\caption{LightGBM-Tuned Hyperparameters Search Space}
\label{tab:lightgbm_hyperparameters}
\begin{tabular}{ll}
\hline
\textbf{Hyperparameter} & \textbf{Search Space} \\ \hline
num\_leaves & $\mathcal{U}(\{5, 6, \dots, 50\})$ \\
max\_depth & $\mathcal{U}(\{3, 4, \dots, 20\})$ \\
learning\_rate & $\log \mathcal{U}(e^{-3}, 1)$ \\
n\_estimators & $\mathcal{U}(\{50, 51, \dots, 2000\})$ \\
min\_child\_weight & $\{10^{-5}, 10^{-3}, 10^{-2}, 10^{-1}, 1, 10, 10^2, 10^3, 10^4\}$ \\
subsample & $\mathcal{U}(0.2, 0.8)$ \\
colsample\_bytree & $\mathcal{U}(0.2, 0.8)$ \\
reg\_alpha & $\{0, 10^{-1}, 1, 2, 5, 7, 10, 50, 100\}$ \\
reg\_lambda & $\{0, 10^{-1}, 1, 5, 10, 20, 50, 100\}$ \\ \hline
\end{tabular}
\end{table}

\subsection{XGBoost}

We use \textit{xgboost} package to run xgboost. Specifically, we use $booster = gbtree$, $early\_stopping\_rounds = 50$, $n\_estimators = 2000$. The hyperparameter search space is shown in table \ref{tab:xgboost_hyperparameters}.

\subsection{LightGBM}

We use \textit{lightgbm} package to run LightGBM with its default implementation. We do hyperparameter search shown in table \ref{tab:lightgbm_hyperparameters} following prior literature \citep{arazi2025tabstar}.

\subsection{Random Forest}

Following \citep{arazi2025tabstar}, we treat Random Forest as a weak baseline to set up a lower-bound reference for each dataset split. We run it with the sklearn package with its default configuration with $n\_estimators = 100$.

\end{document}